\documentclass[runningheads]{llncs}
\usepackage{mathtools}
\usepackage{amsfonts,amssymb}
\usepackage{amsmath}
\allowdisplaybreaks[4]
\usepackage{mathrsfs}

\usepackage{graphicx}
\usepackage{subfigure}
\usepackage[noend]{algpseudocode}
\usepackage{algorithmicx,algorithm}
\usepackage{multirow}
\usepackage{booktabs}
\usepackage{float}
\usepackage{appendix}
\usepackage{marvosym}
\usepackage{ifsym}
\usepackage{url}

\begin{document}
\title{Learning the Implicit Semantic Representation on Graph-Structured Data}
\author{Likang Wu\inst{1} \and
Zhi Li\inst{1} \and Hongke Zhao\inst{2} \and Qi Liu\inst{1} \and Jun Wang\inst{1} \and \\ Mengdi Zhang\inst{3} \and Enhong Chen\inst{1}\textsuperscript{(\Letter)}}
\authorrunning{L. Wu et al.}
\institute{Anhui Province Key Laboratory of Big Data Analysis and Application, University of Science and Technology of China, Hefei, China\\ \email{\{wulk,zhili03\}@mail.ustc.edu.cn}, \email{\{qiliuql,cheneh\}@ustc.edu.cn} \and Tianjin University, Tianjin, China\\ \email{hongke@tju.edu.cn} \and
Meituan-Dianping Group, Beijing, China\\ \email{zhangmengdi02@meituan.com} }
\maketitle              % typeset the header of the contribution
\begin{abstract}
Existing representation learning methods in graph convolutional networks are mainly designed by describing the neighborhood of each node as a perceptual whole, while the implicit semantic associations behind highly complex interactions of graphs are largely unexploited. 
In this paper, we propose a Semantic Graph Convolutional Networks (SGCN) that explores the implicit semantics by learning latent semantic-paths in graphs. In previous work, there are explorations of graph semantics via meta-paths. However, these methods mainly rely on explicit heterogeneous information that is hard to be obtained in a large amount of  graph-structured data. SGCN first breaks through this restriction via leveraging the semantic-paths dynamically and automatically during the node aggregating process. 
To evaluate our idea, we conduct sufficient experiments on several standard datasets, and the empirical results show the superior performance of our model.\footnote{Our code is  available online at \url{https://github.com/WLiK/SGCN_SemanticGCN}}

\keywords{Graph Neural Networks  \and Semantic Representation \and Network Analysis.}
\end{abstract}
\section{Introduction}
The representations of objects (nodes) in large graph-structured data, such as social or biological networks, have been proved extremely effective as feature inputs for graph analysis tasks. Recently, there have been many attempts in the literature to extend neural networks to deal with representation learning of graphs, such as Graph Convolutional Networks (GCN)~\cite{kipf2016semi}, GraphSAGE~\cite{hamilton2017inductive} and Graph Attention Networks (GAT)~\cite{velivckovic2017graph}.

\begin{figure}[t]
   \centering 
    	%% label for first subfigure
    	\includegraphics[width=0.35\columnwidth]{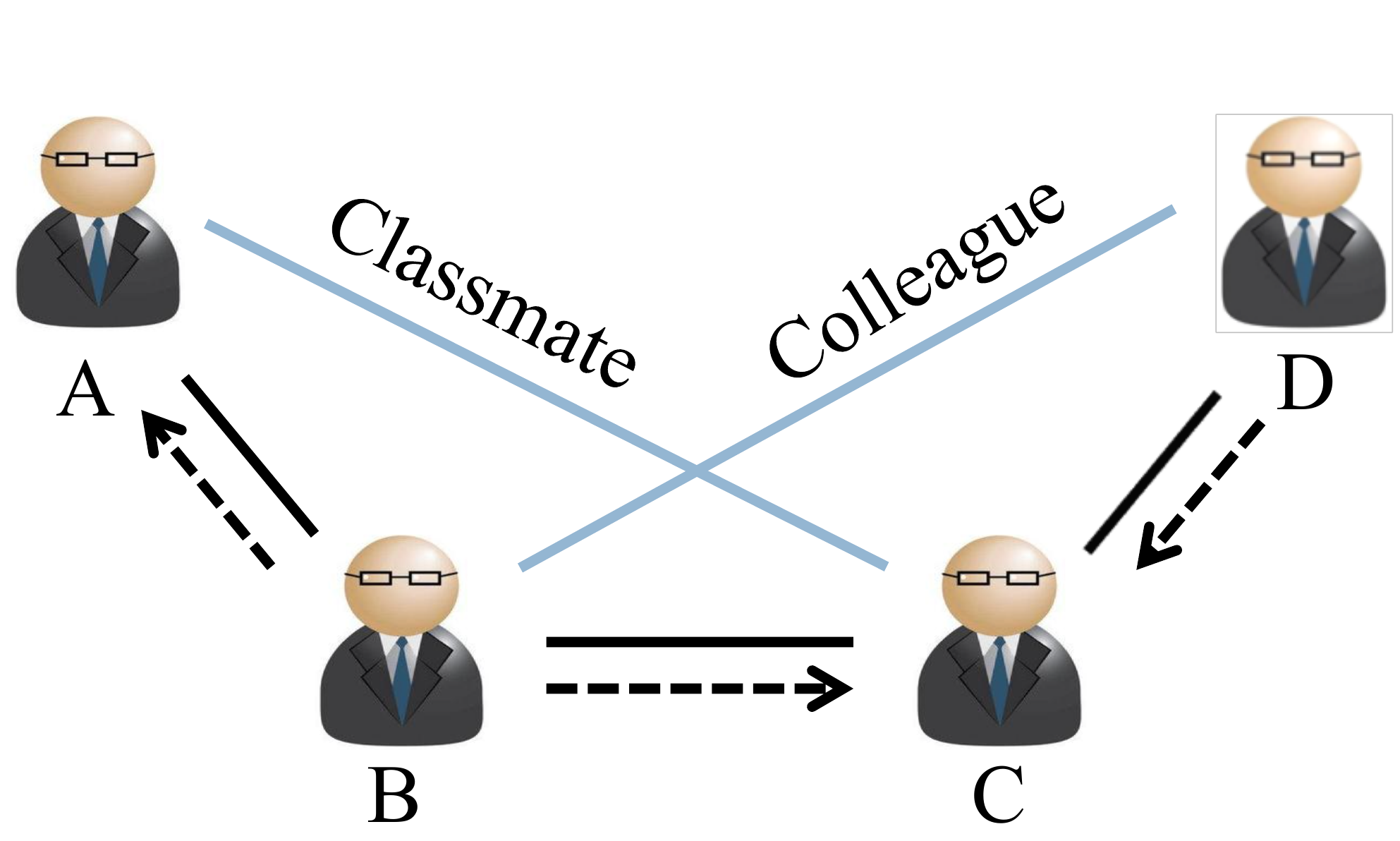}
	%\setlength{\abovecaptionskip}{0.1\columnwidth}
  	%\linespread{1.26}
  	\caption{Example of implicit semantic-paths in a scholar cooperation network. There are not explicit node (relation) types. Behind the same kind of relation (black solid edge), there are implicit factors (dotted line, A is the student of B, B is the advisor of C). So, the path A-B-C expresses ``Student-Advisor-Student'', A and C are ``classmates''. B-C-D expresses ``Advisor-Student-Advisor'', B and D are ``colleagues''. }
    \label{fig:intro} 
\end{figure}

In spite of enormous success, previous graph neural networks mainly proposed representation learning methods by describing the neighborhoods as a perceptual whole, and they have not gone deep into the exploration of semantic information in graphs. Taking the movie network as an example, the paths based on composite relations of ``Movie-Actor-Movie'' and ``Movie-Director-Movie" may reveal two different semantic patterns, i.e., the two movies have the same actor (director). Here the semantic pattern is defined as a specific knowledge expressed by the corresponding path.
Although several researchers~\cite{wang2019heterogeneous,sun2018joint} attempt to capture these graph semantics of composite relations between two objects by meta-paths, existing work relies on the given heterogeneous information such as different types of objects and distinct object connections. However, in the real world, quite a lot of graph-structured data do not have the explicit characteristics. As shown in Figure~\ref{fig:intro}, in a scholar cooperation network, there are usually no explicit node (relation) types and all nodes are connected through the same relation, i.e., ``Co-author''. Fortunately, behind the same relation, there are various implicit factors which may express different connecting reasons, such as ``Classmate'' and ``Colleague'' for the same relation ``Co-author''. These factors can further compose diverse semantic-paths (e.g. ``Student-Advisor-Student'' and ``Advisor-Student-Advisor''), which reveal sophisticated semantic associations and help to generate more informative representations. Then, how to automatically exploit comprehensive semantic patterns based on the implicit factors behind a general graph is a non-trivial problem. 

In general, there are several challenges to solve this problem. Firstly, it is an essential part to adaptively infer latent factors behind graphs. We notice that several researches begin to explore desired latent factors behind a graph by disentangled representations~\cite{ma2019disentangled,liu2019independence}. However, they mainly focus on inferring the latent factors by the disentangled representation learning while failing to discriminatively model the independent implicit factors behind the same connections. Secondly, after discovering the latent factors, how to select the most meaningful semantics and aggregate the diverse semantic information remain largely unexplored. Last but not the least, to further exploit the implicit semantic patterns and to be capable of conducting inductive learning are quite difficult. 

To address above challenges, in this paper, we propose a novel Semantic Graph Convolutional Networks (SGCN), which sheds light on the exploration of implicit semantics in the node aggregating process. Specifically, we first propose a latent factor routing method with the DisenConv layer~\cite{ma2019disentangled} to adaptively infer the probability of each latent factor that may have caused the link from a given node to one of its neighborings. Then, for further exploring the diverse semantic information, we transfer the probability between every two connected nodes to the corresponding semantic adjacent matrix, which can present the semantic-paths in a graph. Afterwards, most semantic strengthen methods like the semantic level attention module can be easily integrated into our model and aggregate the diverse semantic information from these semantic-paths. Finally, to encourage the independence of the implicit semantic factors and conduct the inductive learning, we design an effective joint loss function to maintain the independent mapping channels of different factors. This loss function is able to focus on different semantic characteristics during the training process. 

Specifically, the contributions of this paper can be summarized as follows:

\begin{itemize}
	
	\item We first break the heterogeneous restriction of semantic representations with an end-to-end framework. It automatically infers the independent factor behind the formation of each edge and explores the semantic associations of latent factors behind a graph.
	
	\item We propose a novel Semantic Graph Convolutional Networks (SGCN), to learn node representations by aggregating the implicit semantics from the graph-structured data.
	
	\item We conduct extensive experiments on various real-world graphs datasets to evaluate the performance of the proposed model. The results show the superiority of our proposed model by comparing it with many powerful models.
	
\end{itemize}

\section{Related Works}
Graph neural networks (GNNs)~\cite{gori2005new,scarselli2008graph}, especially graph convolutional networks \cite{henaff2015deep}, have been proven successful in modeling the structured graph data due to its theoretical elegance~\cite{bronstein2017geometric}. They have made new breakthroughs in various tasks, such as node classification~\cite{kipf2016semi} and graph classification~\cite{defferrard2016convolutional}. In the early days, the graph spectral theory~\cite{henaff2015deep} was used to derive a graph convolutional layer. Then, the polynomial spectral filters~\cite{defferrard2016convolutional} greatly reduced the computational cost than before. And, Kipf and Welling~\cite{kipf2016semi} proposed the usage of a linear filter to get further simplification. Along with spectral graph convolution, directly performing graph convolution in the spatial domain was also investigated by many researchers~\cite{duvenaud2015convolutional,hamilton2017inductive}. Among them, graph attention networks~\cite{velivckovic2017graph} has aroused considerable research interest, since it adaptively specify weights to the neighbors of a node by attention mechanism~\cite{bahdanau2014neural,wu2020estimating}.

For semantic learning research, there have been studies explored a kind of semantic-path called meta-path in heterogeneous graph embedding to preserve structural information. ESim~\cite{shang2016meta} learned node representations by searching the user-defined embedding space.  Based on random walk, meta-path2vec~\cite{dong2017metapath2vec} utilized skip-gram to perform a semantic-path. HERec~\cite{shi2018heterogeneous} proposed a type constraint strategy to filter the node sequence and captured the complex semantics reflected in heterogeneous graph. Then, Fan et al.~\cite{fan2018gotcha} suggested a meta-graph2vec model for malware detection, where both the structures and semantics are preserved. Sun et al.~\cite{sun2018joint} proposed meta-graph-based network embedding models, which simultaneously considers the hidden relations of all meta information of a meta-graph. Meanwhile, there were other influential semantic learning approaches in some studies. For instance, many models~\cite{DBLP:journals/jmlr/BleiNJ03,ijcai2020-489,qiao2019structure} were utilized to various fields because of their latent semantic analysis ability.

In heterogeneous graphs, two objects can be connected via different semantic-paths, which are called meta-paths. It depends on the characteristic that this graph structure has different types of nodes and relations. One meta-path $\Phi$ is defined as a path in the form of 
$A_1 \mathop{\longrightarrow}\limits^{R_1} A_2 \mathop{\longrightarrow}\limits^{R_2} \cdot \cdot \cdot \mathop{\longrightarrow}\limits^{R_l} A_{l+1}$ (abbreviated as $A_1 A_2 \cdot \cdot \cdot A_{l+1}$), it describes a composite relation $R = R_1 \circ R_2 \circ \cdot \cdot \cdot \circ R_l$, where $\circ$ denotes the composition operator on relations. Actually, in homogeneous graph, the relationships between nodes are also generated for different reasons (latent factors), so we can implicitly construct various types of relationships to extract various semantic-paths correspond to different semantic patterns, so as to improve the performance of GCN model from the perspective of semantic discovery.

\section{Semantic Graph Convolutional Networks}
In this section, we introduce the Semantic Graph Convolutional Networks (SGCN). We first present the notations, then describe the overall network progressively.

\subsection{Preliminary}
We focus primarily on undirected graphs, and it is straightforward to extend our approach to directed graphs. We define $G = (V,E)$ as a graph, comprised of the nodes set $V$ and edges set $E$, and ${\mid}V{\mid}=N$ denotes the number of nodes. Each node $u \in V$ has a feature vector $\mathbf{x}_u \in  {\mathbb R}^{d_{in}}$. We use $(u, v) \in E$ to indicate that there is an edge between node $u$ and node $v$. Most graph convolutional networks can be regarded as an aggregation function
$f(\cdot)$ that outputs the representations of nodes when given features of each node and its neighbors:
$$\mathbf{y} = f(\mathbf{x}_u, {\mathbf{x}_v : (u, v) \in E} \mid u \in V),$$
where the output $\mathbf{y} \in {\mathbb R}^{N \times d_{out}}$ denotes the representations of nodes. It means that neighborhoods of a node contains rich information, which can be aggregated to describe the node more comprehensively.
Different from previous studies~\cite{kipf2016semi,hamilton2017inductive,velivckovic2017graph}, in our work, proposed $f(\cdot)$ would automatically learn the semantic-path from graph data to explore corresponding semantic pattern.

\subsection{Latent Factor Routing}
Here we aim to introduce the disentangled algorithm that calculates the latent factors between every two objects. We assume that each node is composed of $K$ independent components, hence there are $K$ latent factors to be disentangled. For the node $u \in V$, the hidden representation of $u$ is
$\mathbf{h_u} = [\mathbf{e_{u, 1}}, \mathbf{e_{u, 2}},...,\mathbf{e}_{u, K}] \in  {\mathbb R}^{K \times \frac{d_{out}}{K}}$, where $\mathbf{e}_{u, k} \in  {\mathbb R}^{\frac{d_{out}}{K}} (k =1, 2, ..., K)$ denotes corresponding aspect of node $u$ that is pertinent to the $k$-th disentangled factor. 

In the initial stage, we project its feature vector $\mathbf{x}_u$ into $K$ different subspaces:
\begin{align}
\mathbf{z}_{u, k} = \frac{\sigma{(\mathbf{{W}_{k}} \mathbf{x}_u + \mathbf{b}_k)}}{{\parallel \sigma{(\mathbf{{W}_{k}} \mathbf{x}_u + \mathbf{b}_k)} \parallel}_2} ,
\end{align} 
where $\mathbf{W}_k \in  {\mathbb R}^{d_{in} \times \frac{d_{out}}{K}}$ and $\mathbf{b}_k \in  {\mathbb R}^{\frac{d_{out}}{K}}$ are the mapping parameters and bias of $k$-th subspace, the nonlinear activation function $\sigma$ is $\mathrm{ReLU}$~\cite{nair2010rectified}.
To capture aspect $k$ of node $u$ comprehensively, we construct $\mathbf{e}_{u, k}$ from both $\mathbf{z}_{u, k}$ and $\{\mathbf{z}_{v, k} : (u, v) \in E\}$, which can be utilized to identify the latent factors. Here we learn the probability of each factor by leveraging neighborhood routing mechanism~\cite{ma2019disentangled,liu2019independence}, it is a DisenConv layer: 
\begin{align}
\begin{split}
\mathbf{e}_{u, k}^{t} =  \frac{\mathbf{z}_{u, k} + \sum_{v : (u, v) \in E } \mathbf{p}_{u, v}^{k, t-1} \mathbf{z}_{v, k}}{{\parallel \mathbf{z}_{u, k} + \sum_{v : (u, v) \in E } \mathbf{p}_{u, v}^{k, t-1} \mathbf{z}_{v, k} \parallel}_2},
\end{split}
\end{align}
\begin{align}
\begin{split}
\mathbf{p}_{u, v}^{k, t} =  \frac{\mathrm{exp}(\mathbf{z}_{v, k}^{\top} \mathbf{e}_{u, k}^t)}{\sum_{k=1}^K \mathrm{exp}(\mathbf{z}_{v, k}^{\top} \mathbf{e}_{u, k}^t)},
\end{split}
\end{align}
where iteration $t=1,2,...,T$, $\mathbf{p}_{u, v}^k$ indicates the probability that factor $k$ indicates the reason why node $u$ reaches neighbor $v$, and satisfies $\mathbf{p}_{u, v}^k \geq 0, \sum_{k=1}^{K}\mathbf{p}_{u, v}^k =1$. The neighborhood routing mechanism will iteratively infer $\mathbf{p}_{u, v}^k$ and construct $\mathbf{e}_k$. Note that, there are total $L$ DisenConv layers, $\mathbf{z}_{u, k}$ is assigned the value of $\mathbf{e}_{u, k}^{T}$ finally in each layer $l \leq L-1$, more detail can refer to \textbf{Algorithm} 1.

\subsection{Discriminative Semantic Aggregation}
\begin{figure*}[t]
  \centering
  \subfigure[Multi-graph method]{
    \label{fig:hinmeta} %% label for first subfigure
    \includegraphics[width=0.27\textwidth]{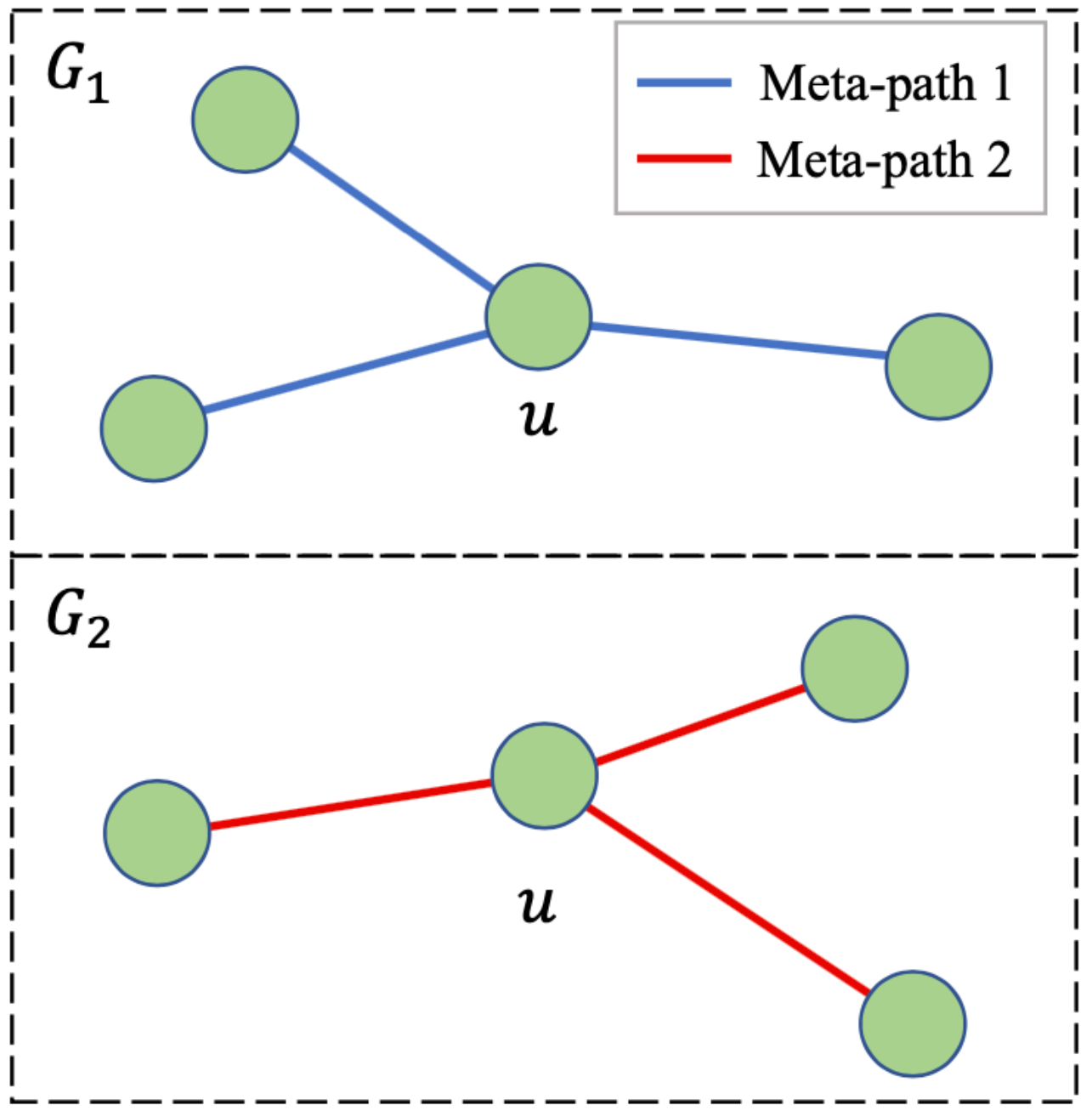}}
  \subfigure[Discriminative semantic aggregation method]{
    \label{fig:homometa} %% label for second subfigure
    \includegraphics[width=0.605\textwidth]{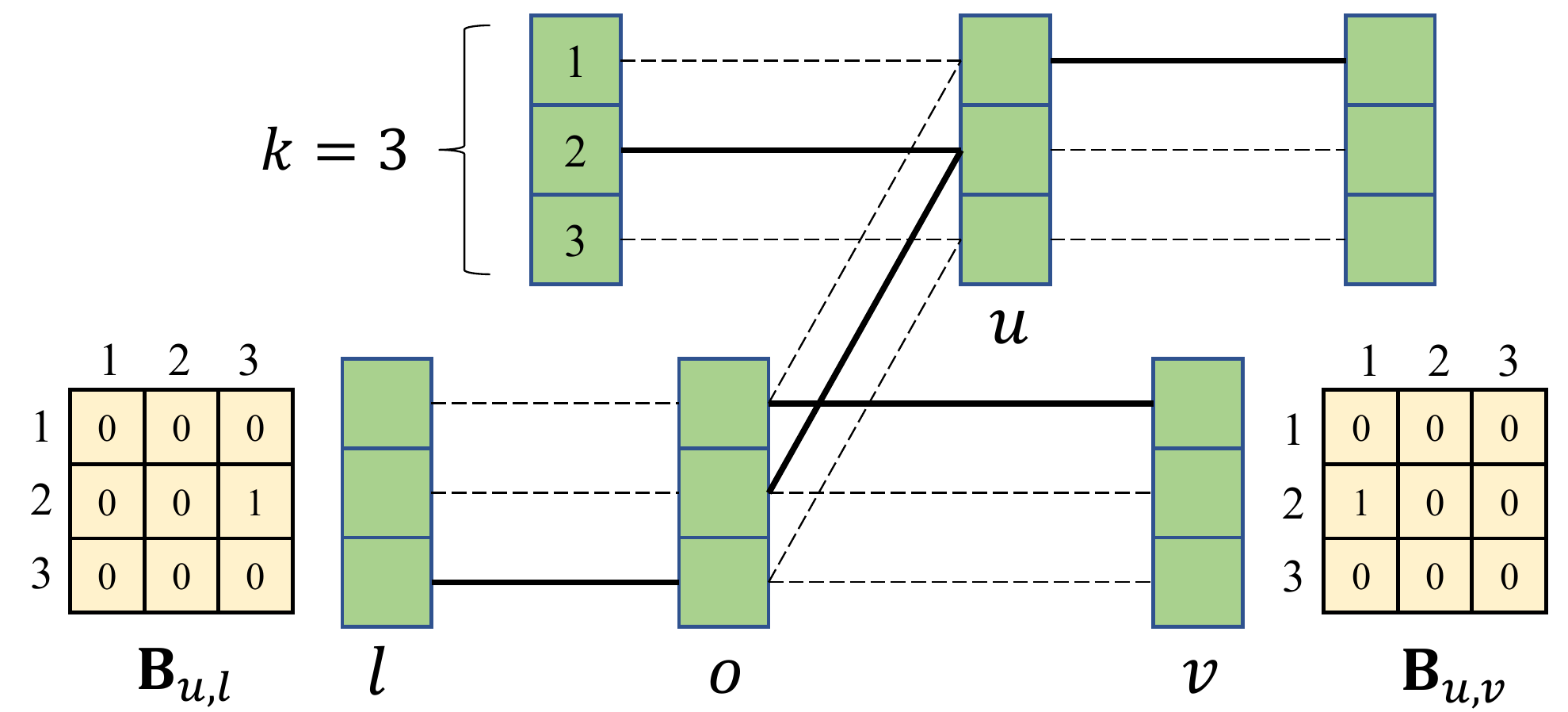}}
  \caption{A previous meta-paths representation on heterogeneous graph and our discriminative semantic aggregation method.}
  \label{fig:metapaths} %% label for entire figure
\end{figure*}

%In this section, we firstly describe a classic method of meta-paths representation based on heterogeneous graph, and then introduce how to ignore the restrictions of heterogeneous conditions in our discriminative semantic aggregation algorithm.
For the data that various relation types between nodes and their corresponding neighbors are explicit and fixed, it is easily to construct multiple sub-semantic graphs as the input data for multiple GCN model. As shown in Figure~\ref{fig:hinmeta} , a heterogeneous graph $G$ contains two different types of meta-paths (meta-path 1, meta-path 2). Then $G$ can be decomposed to multiple graphs $\tilde{G}$ consisting of single semantic graph $G_1$ and $G_2$, where $u$ and its neighbors are connected by path-relation 1(2) for each node $u$ in $G_1(G_2)$.

However, we cannot simply transfer the pre-construct multiple graph method to all network architectures. In detail, for a graph with no different types of edges, we have to judge implicit connecting factors of these edges to find semantic-paths. And the probability of each latent factor is calculated in the iteratively running process as mentioned in last section. To solve this dilemma, we propose a novel algorithm to automatically represent semantic-paths during the model running.

After the latent factor routing process, we get the soft probability matrix of node latents $\mathbf{p} \in {\mathbb R}^{N \times N \times K}$, where $0 \leq \mathbf{p}_{i, j}^{k} \leq 1$ means the possibility that node $i$ connects to $j$ because of the factor $k$. In our model, the latent factor should identify the certain connecting cause of each connected node pair. Here we transfer the probability matrix $\mathbf{p}$ to an semantic adjacent matrix $\mathbf{A}$, so the element in $\mathbf{A}$ only has binary value (0 or 1). In detail, for every node pair $i$ and $j$, $\mathbf{A}_{i, j}^{k}=1$ if $\mathbf{p}_{i, j}^{k}$ denotes the biggest value in $\mathbf{p}_{i, j}$. As shown in Figure~\ref{fig:homometa}, each node is represented by $K$ components. In this graph, every node may connect with others by one relationship from $K$ types, e.g., the relationship between node $u$ and $o$ is $R_2$ (denotes $\mathbf{A}_{u, o}^2 = 1$). 
For node $u$, we can find that it has two semantic-path-based neighbors $l$ and $v$. And, the semantic-paths of $(u, l)$ and $(u, v)$ are two different types which composed by ${\Phi}_{u, o, l} = (\mathbf{A}_{u, o}^{2}, \mathbf{A}_{o, l}^{3}) = R_2 \circ R_3 $ and ${\Phi}_{u, o, v} = (\mathbf{A}_{u, o}^{2}, \mathbf{A}_{o, v}^{1}) = R_2 \circ R_1$ respectively. We define the adjacent matrix $\mathbf{B}$ for virtual semantic-path-based edges, 
\begin{align}
\mathbf{B}_{u, v} = \sum\limits_{[(u, o), (o, v)] \in E}\mathbf{A}_{u, o}^{\top}\mathbf{A}_{o, v}, ~~~\{u, v\} \subset V,
\end{align} 
where $\mathbf{A}_{u, o} \in {\mathbb R}^{K}$, $\mathbf{A}_{o, v} \in {\mathbb R}^{K}$, and $\mathbf{B}_{u, v} \in {\mathbb R}^{K \times K}$. For instance, in Figure~\ref{fig:homometa}, $\mathbf{A}_{u, o} = [0, 1, 0]$, $\mathbf{A}_{o, v} = [1, 0, 0]$, and $\mathbf{A}_{o, l} = [0, 0, 1]$, in this way two semantic-paths start from node $u$ can be expressed as $\mathbf{B}_{u, l}^{2,3} = 1$ and $\mathbf{B}_{u, v}^{2,1} = 1$.

In the semantic information aggregation process, we aggregate the latent vectors connected by corresponding semantic-path as:
\begin{align}
\begin{split}
\mathbf{h}_u &= [\mathbf{e}_{u, 1}, \mathbf{e}_{u, 2}, ... , \mathbf{e}_{u, K}] \in {\mathbb R}^{K \times \frac{d_{out}}{K}},\\
\mathbf{\tilde{h}}_v &= [\mathbf{z}_{v, 1}, \mathbf{z}_{v, 2}, ... , \mathbf{z}_{v, K}] \in {\mathbb R}^{K \times \frac{d_{out}}{K}},\\
%\mathbf{y}_u &= \frac{\mathbf{y}_u + \sum_{v \in \mathbf{V}}\mathbf{B}_{u, v}\mathbf{\tilde{y}}_v}{{\parallel \mathbf{y}_u + \sum_{v \in \mathbf{V}}\mathbf{B}_{u, v}\mathbf{\tilde{y}}_v \parallel}_2}, ~~~u \in \mathbf{V},
\mathbf{y}_u &= \mathbf{h}_u + \mathop{\mathrm{MeanPooling}} \limits_{v \in \mathbf{V}, v \neq u}(\mathbf{B}_{u, v}\mathbf{\tilde{h}}_v), ~~u \in V,\\
%\mathbf{y}_u^{k} &= \mathbf{h}_u^{k}/{\parallel \mathbf{h}_u^{k} \parallel}_2, ~~~1 \leq k \leq K, 
\end{split}
\end{align}
where we just use MeanPooling to avoid large values instead of $\sum\nolimits_{v \in \mathbf{V}}$ operator, and $\mathbf{h}_u, \mathbf{\tilde{h}}_v \in {\mathbb R}^{K \times \frac{d_{out}}{K}}$ are both returned from the last layer of DisenConv operation, in this time that factor probabilities would be stable since the representation of each node considers the influence from neighbors.
According to Eq. (5), the aggregation of two latent representations (end points) of one certain semantic-path denotes the mining result of this semantic relation, e.g., $\mathrm{Pooling}(\mathbf{e}_{u, 2}, \mathbf{z}_{v, 1})$ and $\mathrm{Pooling}(\mathbf{e}_{u, 2}, \mathbf{z}_{l, 3})$ express two different kinds of semantic pattern representations in Figure~\ref{fig:homometa}, $R_2 \circ R_1$ and $R_2 \circ R_3$ respectively. And, for all types of semantic-paths start from node $u$, the weight of each type depends on its frequency. Note that, although the semantic adjacent matrix $\mathbf{A}$ neglects some low probability factors, our semantic paths are integrated with the node states of DisenGCN, which would not lose the crucial information captured by basic GCN model. The advantage of this aggregation method is that our model can distinguish different semantic relations without adding extra parameters, instead of designing various graph convolution networks for different semantic-paths. 
That is to say, the model does not increase the risk of over fitting after the graph semantic-paths learning. Here we only consider 2-order-paths in our model, however, it can be straightly extended to longer path mining.

\subsection{Independence Learning for Mapping Subspaces}
In fact, one type of edge in a meta-path tries to denote one unique meaning, so the $K$ latent factors in our work should not overlap. So, the assumption of using latent factors to construct semantic-paths is that these different factors extracted by latent factor routing module can focus on different connecting causes. In other words, we should encourage the representations of different factors to be of sufficient independence. Before the probability calculating, on our features, the focused point views of $K$ subspaces in Eq. (1) should keep different. Our solution considers that the distance between independence factor representations $\mathbf{z}_{i, k}, k \leq K$ should be sufficient long if they were projected to one subspace.

First, we project the input values $\mathbf{z}$ in Eq. (1) into an unified space to get vectors $\mathbf{Q}$ and  $\mathbf{K}$ as follow:
\begin{align}
\mathbf{Q} = \mathbf{z}\mathbf{w}, \mathbf{K} = \mathbf{z}\mathbf{w},
\end{align}
where $\mathbf{w} \in {\mathbb R}^{\frac{d_{out}}{K} \times \frac{d_{out}}{K}}$ is the projection parameter matrix. Then, the independence loss based on distances between unequal factor representations could be calculated as follow:
\begin{align}
\mathcal{L}_i = \frac{1}{M}\sum \mathrm{softmax}(\frac{\mathbf{Q}\mathbf{K}^{\top}}{\sqrt{\frac{d_{out}}{K}}}) \odot (1 - \mathbf{I}),
\end{align}
where $\mathbf{I} \in {\mathbb R}^{K \times K}$ denotes an identity matrix, $\odot$ is element-wise product, $M = K^2 - K$. Specifically, we learn a lesson from~\cite{vaswani2017attention} that scaling the dot products by $1 /\sqrt{d_{out}/K}$, to counteract the gradients disappear effect for large values. As long as $\mathcal{L}_i$ is minimized in the training process, the distances between different factors tend to be larger, that is, the $K$ subspaces would capture sufficient different information to encourage independence among learned latent factors. 

%\begin{figure*}[t]
%\centering
%\includegraphics[width=.4\textwidth]{subspace.pdf} % Reduce the figure size so that it is slightly narrower than the column.
%\caption{The features in different subspaces keep sufficient independent when the inter-distances of their projections in the unified space are sufficient distinct.}
%\label{fig:subspace}
%\end{figure*}
Next, we would analyze the validity of this optimization.
Latent Factor Routing aims to utilize the disentangled algorithm to calculate the latent factors between every two objects. However, this approach is a variant of von Mises-Fisher (vMF)~\cite{banerjee2005clustering} mixture model, such an EM algorithm cannot optimize the independences of latent factors within the iterative process. And random initialization of the mapping parameters is also not able to promise that subspaces obtain different concerns. For this shortcoming,  we give an assumption: 
\newtheorem{assumption}{Assumption}[section]
\begin{assumption}
The features in different subspaces keep sufficient independent when the margins of their projections in the unified space are sufficiently distinct.
\end{assumption}

This assumption is inspired by the Latent Semantic Analysis algorithm (LSA) ~\cite{landauer1998introduction} that projects multi-dimensional features of a vector space model into a semantic space with less dimensions, which keeps the semantic features of the original space in a statistical sense. So, our optimization approach is listed below:

% \begin{align*}
% \begin{split}
% \mathbf{w} &= \mathop{\arg\min} \sum \mathrm{softmax}(\mathbf{Q}\mathbf{K}^{\mathrm{T}}) \odot (1 - \mathbf{I}),\\
% &=\mathop{\arg\min} \sum\nolimits_u^V \mathrm{softmax}(\mathbf{(z_u w)}\mathbf{(z_u w)}^{\mathrm{T}}) \odot (1 - \mathbf{I}),\\
% %\mathbf{y}_u &= \frac{\mathbf{y}_u + \sum_{v \in \mathbf{V}}\mathbf{B}_{u, v}\mathbf{\tilde{y}}_v}{{\parallel \mathbf{y}_u + \sum_{v \in \mathbf{V}}\mathbf{B}_{u, v}\mathbf{\tilde{y}}_v \parallel}_2}, ~~~u \in \mathbf{V},
% &= \mathop{\arg\min} \sum_{u}^{V}  \frac{\sum_{k_1 \neq k_2} \mathrm{exp}(\mathbf{z}_{u, k_1}\mathbf{w} \cdot \mathbf{z}_{u, k_2}\mathbf{w})}{\sum_{k_1, k_2}\mathrm{exp}(\mathbf{z}_{u, k_1}\mathbf{w} \cdot \mathbf{z}_{u, k_2}\mathbf{w})},\\
% &= \mathop{\arg\max} \sum_{u}^{V} \sum_{k_1 \neq k_2} \mathrm{distance}(\mathbf{z}_{u, k_1}\mathbf{w}, \mathbf{z}_{u, k_2}\mathbf{w}).\\
% S.t.&: ~1 \leq k_1 \leq K, ~1 \leq k_2 \leq K.
% \end{split}
% \end{align*}

\begin{align}
	\mathbf{w} = & \mathop{\arg\min} \sum \mathrm{softmax}(\mathbf{Q}\mathbf{K}^{\mathrm{T}}) \odot (1 - \mathbf{I}), \notag \\
	= & \mathop{\arg\min} \sum\nolimits_u^V \mathrm{softmax}(\mathbf{(z_u w)}\mathbf{(z_u w)}^{\mathrm{T}}) \odot (1 - \mathbf{I}), \notag \\
	= & \mathop{\arg\min} \sum_{u}^{V}  \frac{\sum_{k_1 \neq k_2} \mathrm{exp}(\mathbf{z}_{u, k_1}\mathbf{w} \cdot \mathbf{z}_{u, k_2}\mathbf{w})}{\sum_{k_1, k_2}\mathrm{exp}(\mathbf{z}_{u, k_1}\mathbf{w} \cdot \mathbf{z}_{u, k_2}\mathbf{w})}, \\
	= & \mathop{\arg\max} \sum_{u}^{V} \sum_{k_1 \neq k_2} \mathrm{distance}(\mathbf{z}_{u, k_1}\mathbf{w}, \mathbf{z}_{u, k_2}\mathbf{w}). \notag \\
	S.t.: & ~1 \leq k_1 \leq K, ~1 \leq k_2 \leq K. \notag 
\end{align}

In the above equation, $\mathbf{w}$ denotes the training parameter to be optimized. We ignore the $1/M$ and $1 /\sqrt{d_{out}/K}$ in Eq. (7), because they do not affect the optimization procedure. With the increase of Inter-distances of $K$ subspaces, the IntraVar of factors in each subspace would not larger than the original level (as the random initialization). The InterVar/IntraVar ratio becomes larger, in other word, we get more sufficient independence of mapping subspaces.

\begin{algorithm}[t]
\caption{Semantic Graph Convolutional Networks}
{\bf Input:} 
the feature vector matrix $\mathbf{x} \in {\mathbb R}^{N \times d_{in}}$, the graph $G = (V, E)$, the number of iterations $T$, and the number of disentangle layers $L$.
%\hspace*{0.23in}set of historical projects $\Phi \gets \{{\Phi}_g\ \mid \vee g \in G$\};\\
%static features ${x_i, \vee i \in \Phi \cup G}$;\\
%the funding amounts of first 24 hours ${y_i, \vee i \in \Phi}$;\\
\\
{\bf Output:} 
the representation of node $u$ by $\mathbf{y}_u \in {\mathbb R}^{d_{out}}, \forall u \in V$
\begin{algorithmic}[1]
\For{$i \in V$}
	\For{$k = 1, 2, ..., K$}
	\State $\mathbf{z}_{i, k} \gets \sigma{(\mathbf{{W}_{k}} \mathbf{x}_i + \mathbf{b}_k)} / {\parallel \sigma{(\mathbf{{W}_{k}} \mathbf{x}_i + \mathbf{b}_k)} \parallel}_2 $
	\EndFor{}
\EndFor{}
\State $\mathbf{Q} \gets \mathbf{z}\mathbf{w}_q, \mathbf{K} \gets \mathbf{z}\mathbf{w}_k$
\State $\mathcal{L}_i = \frac{1}{M}\sum \mathrm{softmax}({\mathbf{Q}\mathbf{K}^{\top}}/{\sqrt{\frac{d_{out}}{K}}}) \odot (1 - \mathbf{I})$
\For{disentangle layer $l = 1, 2, ..., L$}
	\State $\mathbf{e}_{u, k}^{t=1} \gets \mathbf{z}_{u, k}, \forall k = 1, 2, ..., K, \forall u \in V$
	\For{routing iteration $t = 1, 2, ..., T$}
		\State Get the soft probability matrix $\mathbf{p}$, where calculating $p_{u, v}^{k, t}$ by Eq. (3)
		\State Update the latent representation $e_{u, k}^{t}, \forall u \in V$ by Eq. (2)
	\EndFor{}
	\State $\mathbf{e}_u \gets \mathrm{dropout}(\mathrm{ReLU}(\mathbf{e}_u)),  \mathbf{z}_{u, k} \gets \mathbf{e}_{u, k}^{t=T},  \forall k = 1, 2, ..., K, \forall u \in V$ ~~$\lhd$ when $l \leq L-1$
%wrong wrong wrong wrong wrong wrong wrong
%这一块公式还有问题，即最后一层的z_uk和e_uk没有区分开
\EndFor{}
\State Transfer $\mathbf{p}$ to hard probability matrix $\mathbf{A}$
\State $\mathbf{B}_{u, v} \gets \sum_{[(u, o), (o, v)] \in E}\mathbf{A}_{u, o}^{\top}\mathbf{A}_{o, v}, ~\{u, v\} \subset V$
\State Get each aggregation $\mathbf{y}_{u}^{k}$ of the latent vectors on semantic-paths by Eq. (5)
%\While{condition}
%????\State ...
%\EndWhile
\State \Return $\{\mathbf{y}_{u}, \forall u \in V\}$, $\mathcal{L}_i$
\end{algorithmic}
\end{algorithm}
\subsection{Algorithm Framework}
In this section, we describe the overall algorithm of SGCN for performing node-related tasks.
For graph $G$, the ground-truth label of node $u$ is $\mathcal{y}_u \in \{0, 1\}^{\mathcal{C}}$, where $\mathcal{C}$ is the number of classes. The details of our algorithm are shown in \textbf{Algorithm} 1. First, we calculate the independence loss $\mathcal{L}_i$ after factor channels capture features. Then, $L$ layers of DisenConv operations would return the stable probability matrix $\mathbf{p}$. After that, the automatic graph semantic-path representation $\mathbf{y}$ is learned based on $\mathbf{p}$. 
To apply $\mathbf{y}$ to different tasks, we design the final layer by a fully-connected layer $\mathbf{y}' = \mathbf{W}_{y} \mathbf{y} + \mathbf{b}_y$, where $\mathbf{W}_{y} \in {\mathbb R}^{d_{out} \times \mathcal{C}}$, $\mathbf{b}_y \in {\mathbb R}^{\mathcal{C}}$. For instance, for the semi-supervised node classification task, we implement 
\begin{align}
\mathcal{L}_s = -\sum_{u \in V^L} \frac{1}{\mathcal{C}} \sum_{c=1}^{\mathcal{C}} \mathcal{y}_u(c)\mathrm{ln} (\hat{\mathbf{y}}_u(c))+\lambda \mathcal{L}_i
\end{align}
 as the loss function, where $\hat{\mathbf{y}}_u = \mathrm{softmax}(\mathbf{y}'_u)$, $V^L$ is the set of labeled nodes, and $\mathcal{L}_i$ would be joint training by sum up with the task loss function. 
For the multi-label classification task, since the label $\mathcal{y}_u$ consists of more than one positive bits, we define the multi-label loss function for node $u$ as: 
\begin{myfont}
\begin{align}
\begin{split}
\mathcal{L}_m = -\frac{1}{\mathcal{C}} \sum_{c=1}^{\mathcal{C}}[\mathcal{y}_u(c) \cdot \mathrm{sigmoid}(\mathbf{y}'_u(c))+ 
(1-\mathcal{y}_u(c)) \cdot \mathrm{sigmoid}(-\mathbf{y}'_u(c))] + \lambda \mathcal{L}_i. 
\end{split}
\end{align}
\end{myfont}
Moreover, for the node clustering task, $\mathbf{y}'$ denotes the input feature of K-Means.

\subsection{Time Complexity Analysis and Optimization}
We should notice a problem in Section 3.3 that the time complexity of Eq. (4-5) by matrix calculation is $O(N  (N-1)  (N-2) K^2 + N ((N-1) K^2 \times \frac{d_{out}}{K} + 2K \frac{d_{out}}{K} )) \approx O(N^3 K^2 + N^2K^2)$. Such a complex time complexity will bring a lot of computing load, so we optimize this algorithm in the actual implementation. For real-world datasets, one node connects to neighbors that are far less than the total number of nodes in the graph. Therefore, when we create the semantic-paths based adjacent matrix, the matrix $\mathbf{\tilde{A}} \in {\mathbb R}^{N \times C \times K}$ is defined to denote 1-order neighbor relationships, $C$ is the maximum number of neighbors that we define, and $\mathbf{\tilde{A}}_{u}^k$ is the id of a neighbor if they are connected by $R_k$, else $\mathbf{\tilde{A}}_{u}^k = 0$. Then the semantic-path relations of type $(R_{k_1}, R_{k_2})$ of $u \in V$ are denoted by $\mathbf{\tilde{B}}_u^{k_1, k_2} = \mathbf{\tilde{A}}[\mathbf{\tilde{A}}[u, :, k_1], :, k_2] \in R^{C\times C} $, and the pooling of this semantic pattern is the mean pooling of $\mathbf{z}[\mathbf{\tilde{B}}_u^{k_1, k_2},~k_2,~: ]$. According to the analysis above, the time complexity can be reduced to $O(K^2(N C^2 + N C^2 \frac{d_{out}}{K})) \approx O(2N K^2 C^2)$. 

\section{Experiments}
In this section, we empirically assess the efficacy of SGCN on several node-related tasks, includes semi-supervised node classification, node clustering and multi-label node classification.  
We then provide node visualization analysis and semantic-paths sampling experiments to verify the validity of our idea. 
\begin{table*}[t]
	\caption{The statistics of datasets.}
	\centering
	\resizebox{.9\textwidth}{!}{
		\smallskip\begin{tabular}{ccccccc}
			\hline
			\textbf{Dataset} & \textbf{Type} & \textbf{Nodes} & \textbf{Edges} & \textbf{Classes} & \textbf{Features} & \textbf{Multi-label} \\
			\hline
			Pubmed & Citation Network & 19,717 & 44,338 & 3 & 500 & False\\
			Citeseer & Citation Network & 3,327 & 4,732 & 6 & 3,703 & False\\
			Cora & Citation Network & 2,708 & 5,429 & 7 & 1,433 & False\\
			Blogcatalog & Social Network & 10,312 & 333,983 & 39 & - & True\\
%			PPI & Biological Network & 3,890 & 76,584 & 50 & - & True\\
			POS & Word Co-occurrence & 4,777 & 184,812 & 40 & - & True\\
			\hline
		\end{tabular}
	}
	\label{table:Datasets}
\end{table*}
\subsection{Experimental Setup}
\subsubsection{Datasets.} 
We conduct our experiments on 5 real-world datasets, Citeseer, Cora, Pubmed, POS and BlogCatalog~\cite{sen2008collective,grover2016node2vec,tang2011leveraging},  whose statistics are listed in Table~\ref{table:Datasets}. The first three citation networks are benchmark datasets for semi-supervised node classification and node clustering.  For graph content, the nodes, edges, and labels in these three represent articles, citations, and research areas, respectively. Their node features correspond a bag-of-words representation of a document. 

POS and BlogCatalog are suitable for multi-label node classification task. Their labels are part-of-speech tags and user interests, respectively. In detail, BlogCatalog is a social relationships network of bloggers who post blogs in the BlogCatalog website. These labels represent the blogger's interests inferred from the text information provided by the blogger. POS (Part-of-Speech) is a co-occurrence network of words appearing in the first million bytes of the Wikipedia dump. The labels in POS denote the Part-of-Speech tags inferred via the Stanford POS-Tagger. 
Due to the two graphs do not provide node features, we use the rows of their adjacency matrices in place of node features for them.

%\begin{table}[t]
%	\caption{Semi-supervised classification (ACC).}
%	\centering
%	\resizebox{.5\columnwidth}{!}{
%       \begin{tabular}[t]{cccc} 
%    		\hline
%		\textbf{Models} & \textbf{Cora} & \textbf{Citeseer} & \textbf{Pubmed} \\
%		\hline
%		MLP & 55.1 & 46.5 & 71.4\\
%%		ManiReg & 59.5 & 60.1 & 70.7\\
%		SemiEmb & 59.0 & 59.6 & 71.1\\
%		LP & 68.0 & 45.3 & 63.0\\
%		DeepWalk & 67.2 & 43.2 & 65.3\\
%		ICA & 75.1 & 69.1 & 73.9\\
%		Planetoid & 75.7 & 64.7 & 77.2\\
%		ChebNet & 81.2 & 69.8 & 74.4\\
%		GCN & 81.5 & 70.3 & 79.0\\
%		MoNet & 81.7 & - & 78.8\\
%		GAT & 83.0 & 72.5 & 79.0\\
%		DisenGCN & 83.7 & 73.4 & 80.5\\
%		IPGDN & 84.1 & 74.0 & 81.2\\
%		\hline
%		SGCN-indep & 84.2 & 73.7 & 82.0\\
%		SGCN-path & 84.6 & \textbf{74.4} & 81.6\\
%		SGCN & \textbf{85.4} & 74.2 & \textbf{82.1}\\
%		\hline
%    	\end{tabular}
%	}
%	\label{table:semi-c}
%\end{table}

\subsubsection{Baselines.}
To demonstrate the advantages of our model, we compare SGCN with some representative graph neural networks, including the graph convolution network (GCN)~\cite{kipf2016semi} and the graph attention network (GAT)~\cite{velivckovic2017graph}. 
In detail, GCN~\cite{kipf2016semi} is a simplified spectral method of node aggregating, while GAT weights a node's neighbors by the attention mechanism. GAT achieves state of the art in many tasks, but it contains far more parameters than GCN and our model. Besides, ChebNet~\cite{defferrard2016convolutional} is a spectral graph convolutional network by means of a Chebyshev expansion of the graph Laplacian, MoNet~\cite{monti2017geometric} extends CNN architectures by learning local, stationary, and compositional task-specific features. And IPGDN~\cite{liu2019independence} is the advanced version of DisenGCN.
%The original implementations of GCN and GAT do not support multi-label tasks. We therefore modify them to use the same multi-label loss function as ours for fair comparison in multi-label tasks.
%We additionally include three node embedding algorithms, including DeepWalk~\cite{perozzi2014deepwalk}, LINE~\cite{tang2015line}, and node2vec~\cite{grover2016node2vec}, for multi-label classification, because they are demonstrated to perform strongly on the multi-label tasks.
We also implement other non-graph convolution network method, including random walk based network embedding DeepWalk~\cite{perozzi2014deepwalk}, link-based classification method ICA~\cite{lu2003link}, inductive embedding based approach Planetoid~\cite{yang2016revisiting}, label propagation approach LP~\cite{zhu2003semi}, semi-supervised embedding learning model SemiEmb~\cite{weston2012deep} and so on.

In addition, we conduct the ablation experiments into nodes classification and clustering to verify the effectiveness of the main components of SGCN: SGCN-path is our complete model without independence loss, and SGCN-indep denotes SGCN without the semantic-path representations.

In the multi-label classification experiment, the original implementations of GCN and GAT do not support multi-label tasks. We therefore modify them to use the same multi-label loss function as ours for fair comparison in multi-label tasks.
We additionally include three node embedding algorithms, including DeepWalk~\cite{perozzi2014deepwalk}, LINE~\cite{tang2015line}, and node2vec~\cite{grover2016node2vec}, because they are demonstrated to perform strongly on the multi-label classification. Besides, we remove IPGDN since it is not designed for multi-label task. 

\subsubsection{Implementation Details.} 
We train our models on one machine with 8 NVIDIA Tesla V100 GPUs. Some experimental results and the settings of common baselines that we follow~\cite{ma2019disentangled,liu2019independence}, and we optimize the parameters of models with Adam~\cite{DBLP:journals/corr/KingmaB14}. Besides, we tune the hyper-parameters of both our model and baselines using hyperopt~\cite{bergstra2013hyperopt}. In detail, for semi-supervised classification and node clustering, we set the number of iterations $T = 6$, the layers $L \in \{ 1, 2, ..., 8 \}$, the number of components $K \in \{ 1, 2,..,7\}$ (denotes the number of mapping channels. Therefore, for our model, the dimension of a component in the SGCN model is $[d_{out} / K] \in \{ 10, 12, ..., 8 \}$), dropout rate $\in \{0.05, 0.10, ..., 0.95\}$, trade-off $\lambda \in \{0.0, 0.5, ..., 10.0\}$, the learning rate $\sim$ loguniform~$[e-8, 1]$, the $l_2$ regularization term $\sim$ loguniform~$[e-10, 1]$. Besides, it should be noted that, in the multi-label node classification,  the output dimension $d_{out}$ is set to 128 to achieve better performance, while setting the dimension of the node embeddings to be 128 as well for other node embedding algorithms. And, when tuning the hyper-parameters, we set the number of components $K \in \{4, 8,...28\}$ in the latent factor routing process.  Here $K = 8$ makes the best result in our experiments.

\subsection{Semi-Supervised Node Classification}
For semi-supervised node classification, there are only 20 labeled instances for each class. It means that the information of neighbors should be leveraged when predicting the labels of target nodes. Here we follow the experimental settings of previous works~\cite{yang2016revisiting,kipf2016semi,velivckovic2017graph}.

We report the classification accuracy (ACC) results in Table 2. The majority of nodes only connect with those neighbors of the same class. According to Table 2, it is obvious that SGCN achieves the best performance amongst all baselines. Here SGCN outperforms the most powerful baseline IPGDN with 1.55\%, 0.47\% and 1.1\% relative accuracy improvements on three datasets, compared with the increasing degrees of previous models, our model express obvious improvements in the node classification task.
And our proposed model achieves the best ACC of 85.4\% on Cora dataset, it is a great improvement on this dataset. On the other hand, in the ablation experiment (the last three rows of Table 2), the complete
\begin{center}
\begin{minipage}[t]{\columnwidth}
\centering
 \begin{minipage}{0.46\columnwidth}
  	\makeatletter\def\@captype{table}\makeatother\caption{Semi-supervised classification.}
	\resizebox{.96\columnwidth}{!}{
       \begin{tabular}[t]{cccc} 
    		\hline
		\textbf{Models} & \textbf{Cora} & \textbf{Citeseer} & \textbf{Pubmed} \\
		\hline
		MLP & 55.1 & 46.5 & 71.4\\
%		ManiReg & 59.5 & 60.1 & 70.7\\
		SemiEmb & 59.0 & 59.6 & 71.1\\
		LP & 68.0 & 45.3 & 63.0\\
		DeepWalk & 67.2 & 43.2 & 65.3\\
		ICA & 75.1 & 69.1 & 73.9\\
		Planetoid & 75.7 & 64.7 & 77.2\\
		ChebNet & 81.2 & 69.8 & 74.4\\
		GCN & 81.5 & 70.3 & 79.0\\
		MoNet & 81.7 & - & 78.8\\
		GAT & 83.0 & 72.5 & 79.0\\
		DisenGCN & 83.7 & 73.4 & 80.5\\
		IPGDN & 84.1 & 74.0 & 81.2\\
		\hline
		SGCN-indep & 84.2 & 73.7 & 82.0\\
		SGCN-path & 84.6 & \textbf{74.4} & 81.6\\
		SGCN & \textbf{85.4} & 74.2 & \textbf{82.1}\\
		\hline
    	\end{tabular}
	}
  \end{minipage}
  \centering
  \begin{minipage}{0.53\columnwidth}
  	\renewcommand\arraystretch{1.2}
        \makeatletter\def\@captype{table}\makeatother\caption{Node clustering with double metrics.}
        \resizebox{.9656\columnwidth}{!}{
         \begin{tabular}[t]{ccccccc}        
          \hline
%		\textbf{Models} & \textbf{Cora} & \textbf{Cora} & \textbf{Citeseer} & \textbf{Citeseer} & \textbf{Pubmed} & \textbf{Pubmed} \\
		\multirow{2}{*}{\textbf{Models}} & \multicolumn{2}{c}{\textbf{Cora}} & \multicolumn{2}{c}{\textbf{Citeseer}} & \multicolumn{2}{c}{\textbf{Pubmed}}\\
		\cline{2-7} & NMI & ARI & NMI & ARI & NMI & ARI\\
		\hline
		SemiEmb & 48.7 & 41.5 & 31.2 & 21.5 & 27.8 &35.2\\
		DeepWalk & 50.3 & 40.8 & 30.5 & 20.6 &29.6 & 36.6\\
		Planetoid & 52.0 & 40.5 & 41.2 & 22.1 & 32.5 & 33.9\\
		ChebNet & 49.8 & 42.4 & 42.6 & 41.5 & 35.6 & 38.6\\
		GCN & 51.7 & 48.9 & 42.8 & 42.8 & 35.0 & 40.9\\
		GAT & 57.0 & 54.1 & 43.1 & 43.6 & 35.0 & 41.4\\
		DIsenGCN & 58.4 & 60.4 & 43.7 & 42.5 & 36.1 & 41.6\\
		IPGDN & 59.2 & 61.0 &44.3 & 43.0 & 37.0 & 42.0\\
		\hline
		SGCN-indep & 60.2 & 59.2 &44.7 & 42.8 & 37.2 & 42.3\\
		SGCN-path & 60.5 & 60.7 &\textbf{45.1} & 44.0 & 37.3 & \textbf{42.8}\\
		SGCN & \textbf{60.7} & \textbf{61.6} &44.9 & \textbf{44.2} & \textbf{37.9} & 42.5\\
		\hline
     	  \end{tabular}
	  }
   \end{minipage}
\end{minipage} 
\end{center}
~\\
SGCN model is superior to either algorithm in at least two datasets. Moreover, we can find that SGCN-indep and SGCN-path are both perform better than previous algorithms  to some degree. It reveals the effectiveness of our semantic-paths mining module and the independence learning for subspaces. 

%\begin{table}[t]
%	\renewcommand\arraystretch{1.15}
%	\caption{Node clustering task with double metrics.}
%	\centering
%	\resizebox{.6\columnwidth}{!}{
%         \begin{tabular}[t]{ccccccc}        
%          \hline
%%		\textbf{Models} & \textbf{Cora} & \textbf{Cora} & \textbf{Citeseer} & \textbf{Citeseer} & \textbf{Pubmed} & \textbf{Pubmed} \\
%		\multirow{2}{*}{\textbf{Models}} & \multicolumn{2}{c}{\textbf{Cora}} & \multicolumn{2}{c}{\textbf{Citeseer}} & \multicolumn{2}{c}{\textbf{Pubmed}}\\
%		\cline{2-7} & NMI & ARI & NMI & ARI & NMI & ARI\\
%		\hline
%		SemiEmb & 48.7 & 41.5 & 31.2 & 21.5 & 27.8 &35.2\\
%		DeepWalk & 50.3 & 40.8 & 30.5 & 20.6 &29.6 & 36.6\\
%		Planetoid & 52.0 & 40.5 & 41.2 & 22.1 & 32.5 & 33.9\\
%		ChebNet & 49.8 & 42.4 & 42.6 & 41.5 & 35.6 & 38.6\\
%		GCN & 51.7 & 48.9 & 42.8 & 42.8 & 35.0 & 40.9\\
%		GAT & 57.0 & 54.1 & 43.1 & 43.6 & 35.0 & 41.4\\
%		DIsenGCN & 58.4 & 60.4 & 43.7 & 42.5 & 36.1 & 41.6\\
%		IPGDN & 59.2 & 61.0 &44.3 & 43.0 & 37.0 & 42.0\\
%		\hline
%		SGCN-indep & 60.2 & 59.2 &44.7 & 42.8 & 37.2 & 42.3\\
%		SGCN-path & 60.5 & 60.7 &\textbf{45.1} & 44.0 & 37.3 & \textbf{42.8}\\
%		SGCN & \textbf{60.7} & \textbf{61.6} &44.9 & \textbf{44.2} & \textbf{37.9} & 42.5\\
%		\hline
%     	  \end{tabular}
%	  }
%	\label{table:cluster}
%\end{table}
\subsection{Multi-label Node Classification}
\begin{figure}[t]
    \centering 
  	\subfigure[Macro-F1 POS]{
    	\label{fig:macropos} %% label for first subfigure
    	\includegraphics[width=0.35\columnwidth]{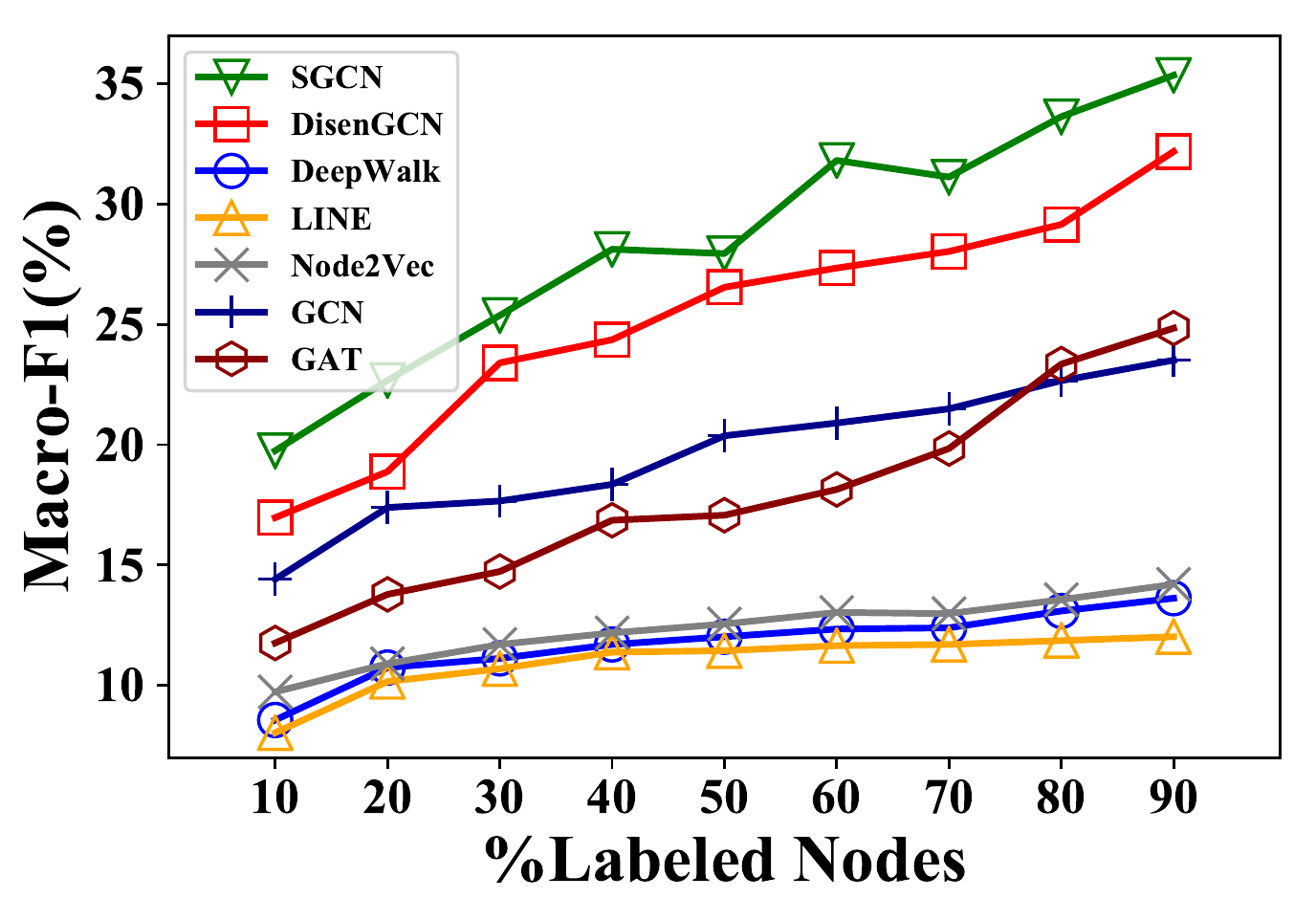}}
%  	\subfigure[Macro-F1 PPI]{
%    	\label{fig:macroppi} %% label for second subfigure
%    	\includegraphics[width=0.32\columnwidth]{Macro_F1_POS}}
	\subfigure[Macro-F1 Blogcatalog]{
    	\label{fig:macroblog} %% label for second subfigure
    	\includegraphics[width=0.35\columnwidth]{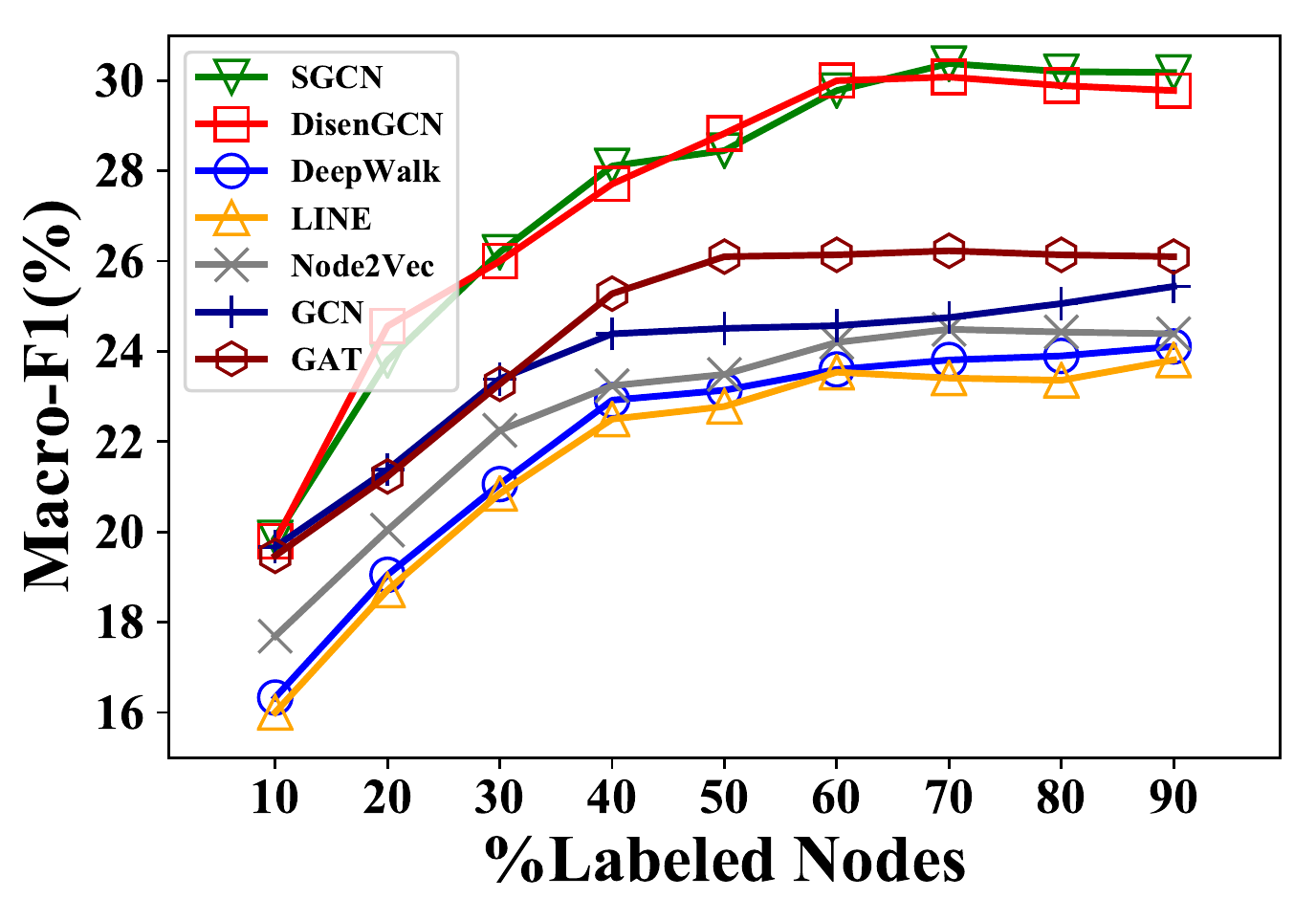}}
	 \subfigure[Micro-F1 POS]{
    	\label{fig:micropos} %% label for first subfigure
    	\includegraphics[width=0.35\columnwidth]{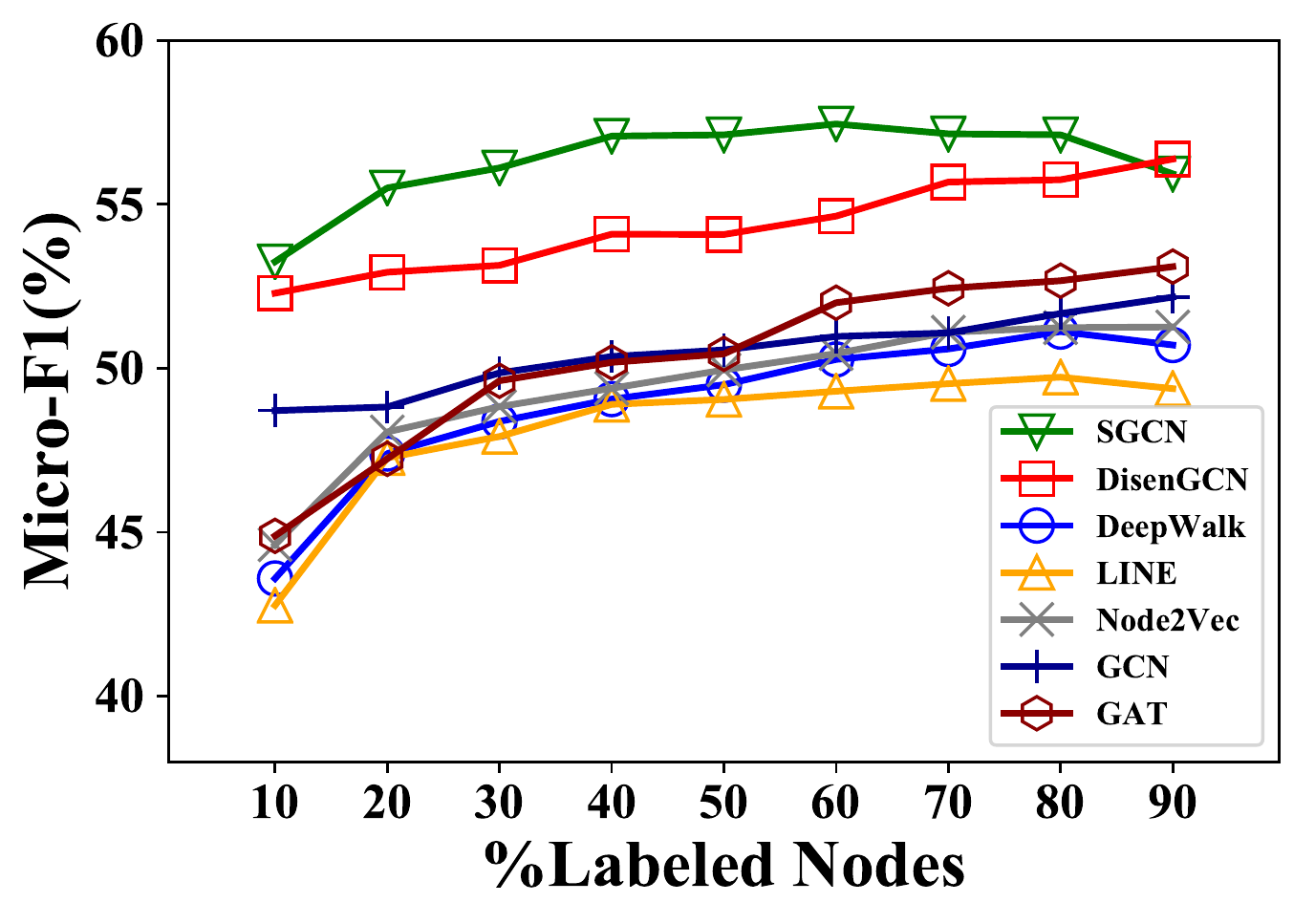}}
%  	\subfigure[Micro-F1 PPI]{
%    	\label{fig:microppi} %% label for second subfigure
%    	\includegraphics[width=0.32\columnwidth]{Micro_F1_POS}}
	\subfigure[Micro-F1 Blogcatalog]{
    	\label{fig:microblog} %% label for second subfigure
    	\includegraphics[width=0.35\columnwidth]{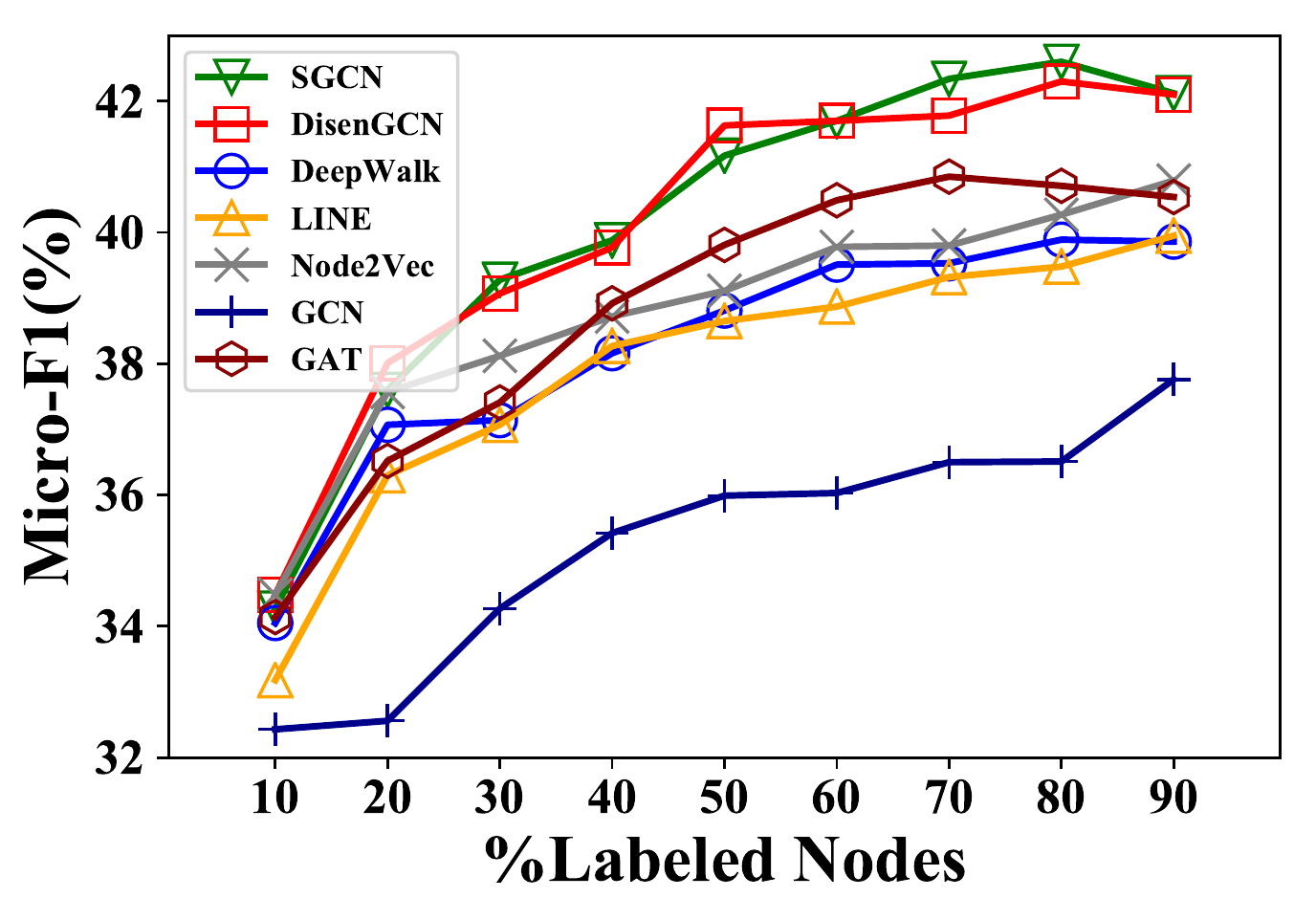}}
  	\caption{ Results of multi-label node classification.}
    \label{fig:multiclass} 
\end{figure}

In the multi-label classification experiment, every node is assigned one or more labels from a finite set $\mathcal{L}$. We follow node2vec~\cite{grover2016node2vec} and report the performance of each method while varying the number of nodes labeled for training from 10\% $|V|$ to 90\% $ |V|$, where $|V|$ is the total number of nodes. The rest of nodes are split equally to form a validation set and a test set. Then with the best hyper-parameters on the validation sets, we report the averaged performance of 30 runs on each multi-label test set. Here we summarize the results of multi-label node classification by Macro-F1 and Micro-F1 scores in Figure~\ref{fig:multiclass}.
Firstly, there is an obvious point that proposed SGCN model achieves the best performances in both two datasets. 
Compared with DisenGCN model, SGCN combines with semantic semantic-paths can achieve the biggest improvement of 20.0\% when we set 10\% of labeled nodes in POS dataset. The reason may be that the relation type of POS dataset is Word Co-occurrence, there are lots of regular explicit or implicit semantics amongst these relationships between different words. In the other dataset, although SGCN does not show a full lead but achieves the highest accuracy on both indicators. We find that the GCN-based algorithms are usually superior to the traditional node embedding algorithms in overall effect. Although for the Micro-F1 score on Blogcatalog, GCN produces the poor results. In addition, the SGCN algorithm can make both Macro-F1 and Micro-F2 achieve good results at the same time, and there will be no bad phenomenon in one of them. Because this approach would not ignore the information provided by the classes with few samples but important semantic relationships.

\subsection{Node Clustering}
To further evaluate the embeddings learned from the above algorithms, we also conduct the clustering task. Following~\cite{liu2019independence}, for our model and each baseline, we obtain its node embedding via feed forward when the model is trained. Then we input the node embedding to the K-Means algorithm to cluster nodes. The ground-truth is the same as that of node classification task, and the number of clusters $K$ is set to the number of classes. In detail, we employ two metrics of Normalized Mutual Information (NMI) and Average Rand Index (ARI) to validate the clustering results. Since the performance of K-Means is affected by initial centroids, we repeat the process for 20 times and report the average results in Table 3.
As can be seen in Table 3, SGCN consistently outperforms all baselines, and GNN-based algorithms usually achieve better performance. Besides, with the semantic-path representation, SGCN and SGCN-path performs significantly better than DisenGCN and IPGDN, our proposed algorithm gets the best results on both NMI and ARI. It shows that SGCN captures a more meaningful node embedding via learning semantic patterns from graph.

% \begin{figure}[t]
%     \centering 
%   	\subfigure[DisenGCN]{
%     	\label{fig:tsne1} %% label for first subfigure
%     	\includegraphics[width=0.35\columnwidth]{main_tsne}}
%   	\subfigure[SGCN]{
%     	\label{fig:tsne2} %% label for second subfigure
%     	\includegraphics[width=0.35\columnwidth]{sameta_main_tsne_3}}
% 	%\setlength{\abovecaptionskip}{0.001\columnwidth}
%   	\caption{ Node representation visualization of Cora.}
%     \label{fig:visual} 
% \end{figure}
\begin{figure}[t]
  \begin{minipage}{0.61\columnwidth} %
  \vspace{0.13cm}
    \centering 
  	\subfigure[DisenGCN]{
    	\label{fig:tsne1} %% label for first subfigure
    	\includegraphics[width=0.46\columnwidth]{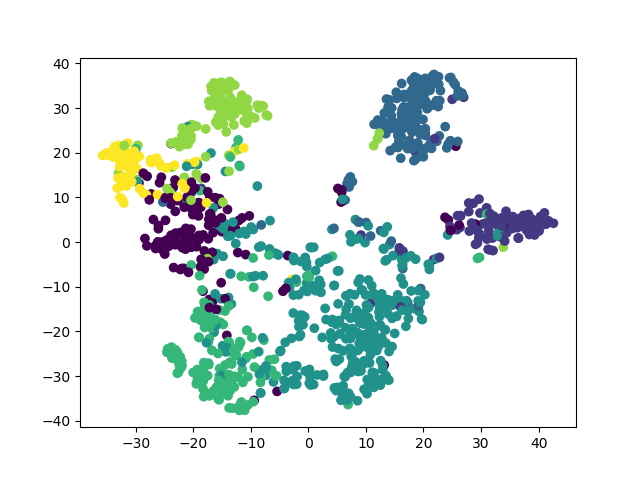}}
  	\subfigure[SGCN]{
    	\label{fig:tsne2} %% label for second subfigure
    	\includegraphics[width=0.46\columnwidth]{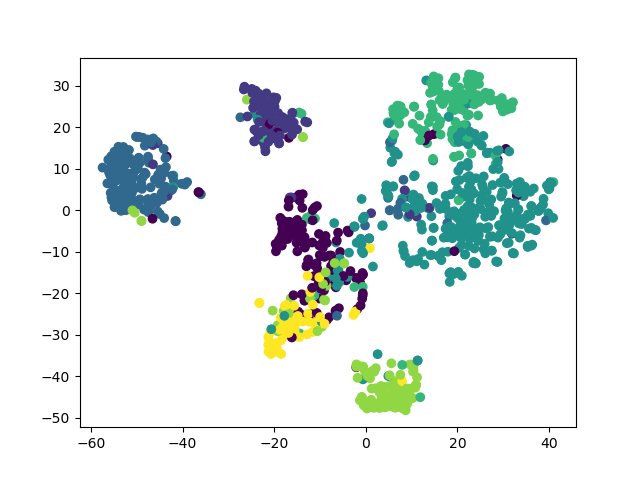}}
  	\caption{Node representation visualization of Cora.}
    \label{fig:side:a} 
  \end{minipage}% 
  \begin{minipage}{0.39\columnwidth} 
  \vspace{-0.02cm}
   \centering 
       \subfigure{
    	\label{fig:tsne1} %% label for first subfigure
    	\includegraphics[width=0.82\columnwidth]{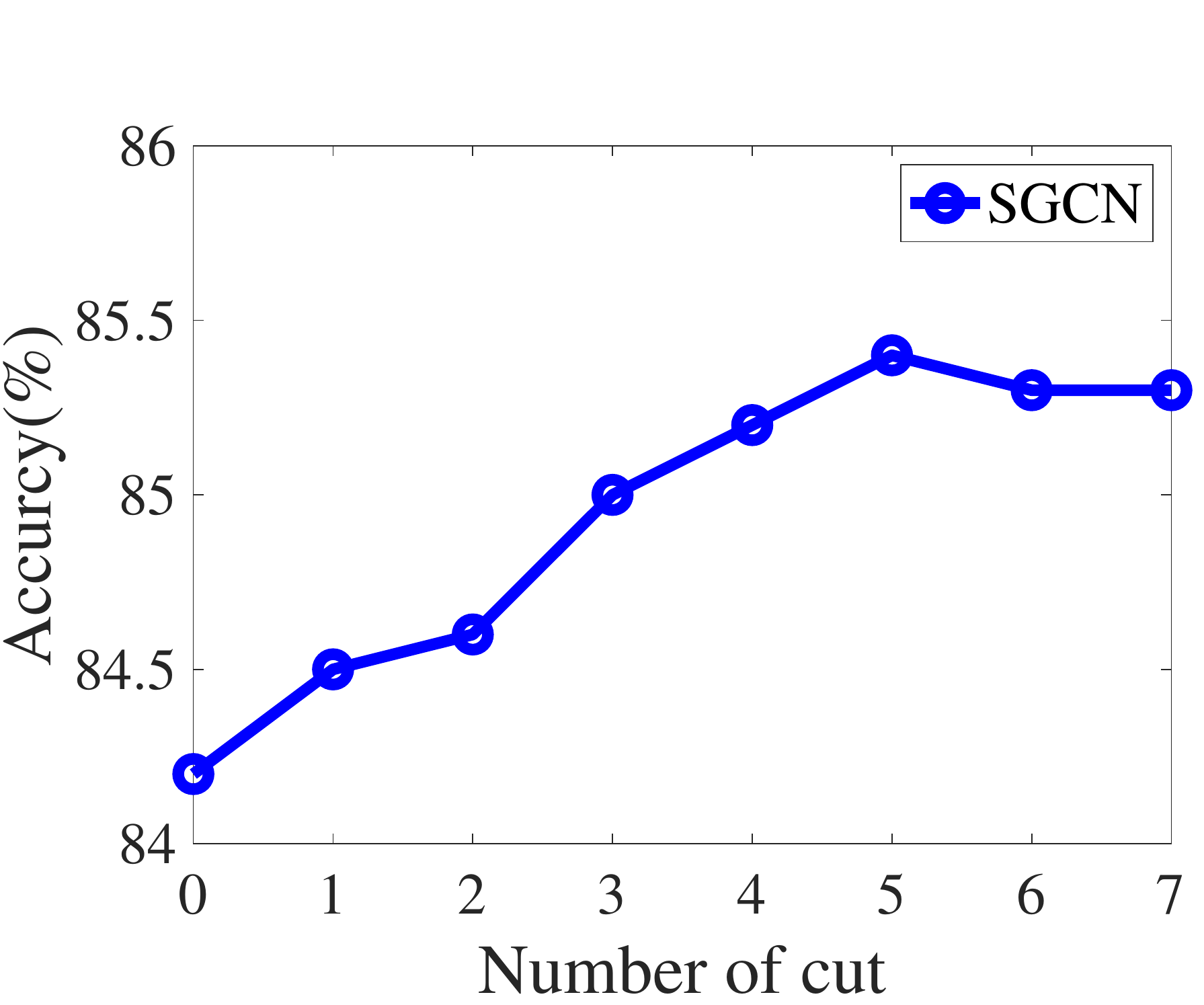}}
% 	\setlength{\abovecaptionskip}{0.1\columnwidth}
  	%\linespread{1.26}
  	\vspace{-0.08cm}
  	\caption{Semantic-paths sampling.}
    \label{fig:side:b} 
  \end{minipage} 
\end{figure}

%\subsection{Visualization Analysis and Semantic-paths Sampling}
\subsection{Visualization Analysis and Semantic-paths Sampling}
We try to demonstrate the intuitive changes of node representations after incorporating semantic patterns. Therefore, we utilize t-SNE~\cite{maaten2008visualizing} to transform feature representations (node embedding) of SGCN and DisenGCN into a 2-dimensional space to make a more intuitive visualization. Here we visualize the node embedding of Cora (actually, the change of representation visualization is similar in other datasets), where different colors denote different research areas.
According to Figure~\ref{fig:side:a}, there is a phenomenon that the visualization of SGCN is more distinguishable than DisenGCN. It demonstrates that the embedding learned by SGCN presents a high intra-class similarity and separates papers into different research areas with distinct boundaries. On the contrary, DisenGCN dose not perform well since the inter-margin of clusters are not distinguishable enough. In several clusters, many nodes belong to different areas are mixed with others.

Then, to explore the influence of different scales of semantic-paths on our model performance, we implement a semantic-paths sampling experiment on Cora. As mentioned in the section 3.6, for capturing different numbers of semantic paths, we change the hyper-parameter of cut size $C$ to restrict the sampling size on each node's neighbors. As shown in Figure~\ref{fig:side:b}, the SGCN model with the path representation achieves higher performances than the first point ($C = 0$).  From the perspective of global trend, with the increase of $C$, the classification accuracy of SGCN model is also improved steady, although it get the highest score when $C = 5$. It means that GCN model combines with more sufficient scale semantic-paths can really learn better node representations.

\section{Conclusion}
In this paper, we proposed a novel framework named Semantic Graph Convolutional Networks which incorporates the semantic-paths automatically during the node aggregating process. Therefore, SGCN provided the semantic learning ability to general graph algorithms. We conducted extensive experiments on various real-world datasets to evaluate the superior performance of our proposed model.
Moreover, our method has good expansibility, all kinds of path-based algorithms in the graph embedding field can be directly applied in SGCN to adapt to different tasks, we will take more explorations in future work. 

\section{Acknowledgements}
This research was partially supported by grants from the National Key Research and Development Program of China (No. 2018YFC0832101), and the National Natural Science Foundation of China (No.s U20A20229 and 61922073). This research was also supported by Meituan-Dianping Group.
 \bibliographystyle{splncs04}
 \bibliography{neurips_2019}

\begin{thebibliography}{10}
\providecommand{\url}[1]{\texttt{#1}}
\providecommand{\urlprefix}{URL }
\providecommand{\doi}[1]{https://doi.org/#1}

\bibitem{bahdanau2014neural}
Bahdanau, D., Cho, K., Bengio, Y.: Neural machine translation by jointly
  learning to align and translate. arXiv preprint arXiv:1409.0473  (2014)

\bibitem{banerjee2005clustering}
Banerjee, A., Dhillon, I.S., Ghosh, J., Sra, S.: Clustering on the unit
  hypersphere using von mises-fisher distributions. J. Mach. Learn. Res.
  \textbf{6}(Sep),  1345--1382 (2005)

\bibitem{bergstra2013hyperopt}
Bergstra, J., Yamins, D., Cox, D.D.: Hyperopt: A python library for optimizing
  the hyperparameters of machine learning algorithms. In: Proceedings of the
  12th Python in science conference. pp. 13--20. Citeseer (2013)

\bibitem{DBLP:journals/jmlr/BleiNJ03}
Blei, D.M., Ng, A.Y., Jordan, M.I.: Latent dirichlet allocation. J. Mach.
  Learn. Res.  \textbf{3},  993--1022 (2003),
  \url{http://jmlr.org/papers/v3/blei03a.html}

\bibitem{bronstein2017geometric}
Bronstein, M.M., Bruna, J., LeCun, Y., Szlam, A., Vandergheynst, P.: Geometric
  deep learning: going beyond euclidean data. IEEE Signal Processing Magazine
  \textbf{34}(4),  18--42 (2017)

\bibitem{defferrard2016convolutional}
Defferrard, M., Bresson, X., Vandergheynst, P.: Convolutional neural networks
  on graphs with fast localized spectral filtering. In: Advances in neural
  information processing systems. pp. 3844--3852 (2016)

\bibitem{dong2017metapath2vec}
Dong, Y., Chawla, N.V., Swami, A.: metapath2vec: Scalable representation
  learning for heterogeneous networks. In: Proceedings of the 23rd ACM SIGKDD
  international conference on knowledge discovery and data mining. pp. 135--144
  (2017)

\bibitem{duvenaud2015convolutional}
Duvenaud, D.K., Maclaurin, D., Iparraguirre, J., Bombarell, R., Hirzel, T.,
  Aspuru-Guzik, A.: Convolutional networks on graphs for learning molecular
  fingerprints. In: Advances in neural information processing systems. pp.
  2224--2232 (2015)

\bibitem{fan2018gotcha}
Fan, Y., Hou, S., Zhang, Y., Ye, Y., Abdulhayoglu, M.: Gotcha-sly malware!
  scorpion a metagraph2vec based malware detection system. In: Proceedings of
  the 24th ACM SIGKDD. pp. 253--262 (2018)

\bibitem{gori2005new}
Gori, M., Monfardini, G., Scarselli, F.: A new model for learning in graph
  domains. In: Proceedings. 2005 IEEE International Joint Conference on Neural
  Networks, 2005. vol.~2, pp. 729--734. IEEE (2005)

\bibitem{grover2016node2vec}
Grover, A., Leskovec, J.: node2vec: Scalable feature learning for networks. In:
  Proceedings of the 22nd ACM SIGKDD. pp. 855--864 (2016)

\bibitem{hamilton2017inductive}
Hamilton, W., Ying, Z., Leskovec, J.: Inductive representation learning on
  large graphs. In: NIPS. pp. 1024--1034 (2017)

\bibitem{henaff2015deep}
Henaff, M., Bruna, J., LeCun, Y.: Deep convolutional networks on
  graph-structured data. arXiv preprint arXiv:1506.05163  (2015)

\bibitem{DBLP:journals/corr/KingmaB14}
Kingma, D.P., Ba, J.: Adam: {A} method for stochastic optimization. In: 3rd
  International Conference on Learning Representations, {ICLR} 2015, San Diego,
  CA, USA, May 7-9, 2015, Conference Track Proceedings (2015)

\bibitem{kipf2016semi}
Kipf, T.N., Welling, M.: Semi-supervised classification with graph
  convolutional networks. arXiv preprint arXiv:1609.02907  (2016)

\bibitem{landauer1998introduction}
Landauer, T.K., Foltz, P.W., Laham, D.: An introduction to latent semantic
  analysis. Discourse processes  \textbf{25}(2-3),  259--284 (1998)

\bibitem{ijcai2020-489}
Li, Z., Wu, B., Liu, Q., Wu, L., Zhao, H., Mei, T.: Learning the compositional
  visual coherence for complementary recommendations. In: IJCAI-20. pp.
  3536--3543

\bibitem{liu2019independence}
Liu, Y., Wang, X., Wu, S., Xiao, Z.: Independence promoted graph disentangled
  networks. Proceedings of the AAAI Conference on Artificial Intelligence
  (2020)

\bibitem{lu2003link}
Lu, Q., Getoor, L.: Link-based classification. In: Proceedings of the 20th
  International Conference on Machine Learning (ICML-03). pp. 496--503 (2003)

\bibitem{ma2019disentangled}
Ma, J., Cui, P., Kuang, K., Wang, X., Zhu, W.: Disentangled graph convolutional
  networks. In: International Conference on Machine Learning. pp. 4212--4221
  (2019)

\bibitem{maaten2008visualizing}
Maaten, L.v.d., Hinton, G.: Visualizing data using t-sne. Journal of machine
  learning research  \textbf{9}(Nov),  2579--2605 (2008)

\bibitem{monti2017geometric}
Monti, F., Boscaini, D., Masci, J., Rodola, E., Svoboda, J., Bronstein, M.M.:
  Geometric deep learning on graphs and manifolds using mixture model cnns. In:
  IEEE Conference on Computer Vision and Pattern Recognition. pp. 5115--5124
  (2017)

\bibitem{nair2010rectified}
Nair, V., Hinton, G.E.: Rectified linear units improve restricted boltzmann
  machines. In: Proceedings of the 27th international conference on machine
  learning (ICML-10). pp. 807--814 (2010)

\bibitem{perozzi2014deepwalk}
Perozzi, B., Al-Rfou, R., Skiena, S.: Deepwalk: Online learning of social
  representations. In: Proceedings of the 20th ACM SIGKDD. pp. 701--710 (2014)

\bibitem{qiao2019structure}
Qiao, L., Zhao, H., Huang, X., Li, K., Chen, E.: A structure-enriched neural
  network for network embedding. Expert Systems with Applications pp. 300--311
  (2019)

\bibitem{scarselli2008graph}
Scarselli, F., Gori, M., Tsoi, A.C., Hagenbuchner, M., Monfardini, G.: The
  graph neural network model. IEEE Transactions on Neural Networks
  \textbf{20}(1),  61--80 (2008)

\bibitem{sen2008collective}
Sen, P., Namata, G., Bilgic, M., Getoor, L., Galligher, B., Eliassi-Rad, T.:
  Collective classification in network data. AI magazine  \textbf{29}(3),
  93--93 (2008)

\bibitem{shang2016meta}
Shang, J., Qu, M., Liu, J., Kaplan, L.M., Han, J., Peng, J.: Meta-path guided
  embedding for similarity search in large-scale heterogeneous information
  networks. arXiv preprint arXiv:1610.09769  (2016)

\bibitem{shi2018heterogeneous}
Shi, C., Hu, B., Zhao, W.X., Philip, S.Y.: Heterogeneous information network
  embedding for recommendation. IEEE Transactions on Knowledge and Data
  Engineering  \textbf{31}(2),  357--370 (2018)

\bibitem{sun2018joint}
Sun, L., He, L., Huang, Z., Cao, B., Xia, C., Wei, X., Philip, S.Y.: Joint
  embedding of meta-path and meta-graph for heterogeneous information networks.
  In: 2018 IEEE International Conference on Big Knowledge. pp. 131--138. IEEE
  (2018)

\bibitem{tang2015line}
Tang, J., Qu, M., Wang, M., Zhang, M., Yan, J., Mei, Q.: Line: Large-scale
  information network embedding. In: Proceedings of the 24th international
  conference on world wide web. pp. 1067--1077 (2015)

\bibitem{tang2011leveraging}
Tang, L., Liu, H.: Leveraging social media networks for classification. Data
  Mining and Knowledge Discovery  \textbf{23}(3),  447--478 (2011)

\bibitem{vaswani2017attention}
Vaswani, A., Shazeer, N., Parmar, N., Uszkoreit, J., Jones, L., Gomez, A.N.,
  Kaiser, {\L}., Polosukhin, I.: Attention is all you need. In: Advances in
  neural information processing systems. pp. 5998--6008 (2017)

\bibitem{velivckovic2017graph}
Veli{\v{c}}kovi{\'c}, P., Cucurull, G., Casanova, A., Romero, A., Lio, P.,
  Bengio, Y.: Graph attention networks. arXiv preprint arXiv:1710.10903  (2017)

\bibitem{wang2019heterogeneous}
Wang, X., Ji, H., Shi, C., Wang, B., Ye, Y., Cui, P., Yu, P.S.: Heterogeneous
  graph attention network. In: The World Wide Web Conference. pp. 2022--2032
  (2019)

\bibitem{weston2012deep}
Weston, J., Ratle, F., Mobahi, H., Collobert, R.: Deep learning via
  semi-supervised embedding. In: Neural networks: Tricks of the trade, pp.
  639--655. Springer (2012)

\bibitem{wu2020estimating}
Wu, L., Li, Z., Zhao, H., Pan, Z., Liu, Q., Chen, E.: Estimating early
  fundraising performance of innovations via graph-based market environment
  model. In: AAAI. pp. 6396--6403 (2020)

\bibitem{yang2016revisiting}
Yang, Z., Cohen, W.W., Salakhutdinov, R.: Revisiting semi-supervised learning
  with graph embeddings. arXiv preprint arXiv:1603.08861  (2016)

\bibitem{zhu2003semi}
Zhu, X., Ghahramani, Z., Lafferty, J.D.: Semi-supervised learning using
  gaussian fields and harmonic functions. In: Proceedings of the 20th
  International conference on Machine learning (ICML-03). pp. 912--919 (2003)

\end{thebibliography}

\end{document}